Disorders of the Nervous System

# Clonal Analysis of Newborn Hippocampal Dentate Granule Cell Proliferation and Development in Temporal Lobe Epilepsy[1,2,3]

Shatrunjai P. Singh,[1,2] Candi L. LaSarge,[1] Amen An,[1,4] John J. McAuliffe,[1] and Steve C. Danzer[1,2,3]

## Abstract

Hippocampal dentate granule cells are among the few neuronal cell types generated throughout adult life in mammals. In the normal brain, new granule cells are generated from progenitors in the subgranular zone and integrate in a typical fashion. During the development of epilepsy, granule cell integration is profoundly altered. The new cells migrate to ectopic locations and develop misoriented "basal" dendrites. Although it has been established that these abnormal cells are newly generated, it is not known whether they arise ubiquitously throughout the progenitor cell pool or are derived from a smaller number of "bad actor" progenitors. To explore this question, we conducted a clonal analysis study in mice expressing the Brainbow fluorescent protein reporter construct in dentate granule cell progenitors. Mice were examined 2 months after pilocarpine-induced status epilepticus, a treatment that leads to the development of epilepsy. Brain sections were rendered translucent so that entire hippocampi could be reconstructed and all fluorescently labeled cells identified. Our findings reveal that a small number of progenitors produce the majority of ectopic cells following status epilepticus, indicating that either the affected progenitors or their local microenvironments have become pathological. By contrast, granule cells with "basal" dendrites were equally distributed among clonal groups. This indicates that these progenitors can produce normal cells and suggests that global factors sporadically disrupt the dendritic development of some new cells. Together, these findings strongly predict that distinct mechanisms regulate different aspects of granule cell pathology in epilepsy.

### Significance Statement

Epileptogenic injuries disrupt adult neurogenesis, leading to the abnormal integration of adult-generated granule cells. The newborn cells exhibit a variety of pathologies, including dendritic abnormalities and migration defects. It was not known, however, whether all progenitors contributed equally to the accumulation of these abnormal cells or whether a distinct subset of progenitors was responsible. Here, we performed a clonal analysis study of progenitor cell activity following status epilepticus. Our results reveal that a small subset of progenitors produces the majority of ectopic granule cells, while cells with abnormal dendrites arise ubiquitously throughout the progenitor pool. Together, these findings demonstrate a newly understood complexity among progenitors in producing abnormal granule cells in epilepsy.

## Introduction

Hippocampal dentate granule cells (DGCs) are generated throughout life from progenitor cells located in the subgranular zone, a proliferative region located between the granule cell body layer and the hilus. A subset of progenitor cells in this region expresses the Gli1 transcription factor, a Krüppel family zinc finger protein activated by the sonic hedgehog signal transduction cascade (Ahn and Joyner, 2005; Palma et al., 2005). Sonic hedgehog is a key regulator of cell proliferation (Lai et al., 2003). Gli1-expressing type 1 progenitor cells are morphologically characterized by the presence of a radial process that projects into the dentate inner molecular layer. They exhibit the stem cell characteristics of self-renewal and multipotency, and give rise to intermediate progenitors (type 2 cells; transient amplifying cells), which, in turn, give rise to postnatally generated dentate granule cells. Type 1 cells can also give rise to astrocytes (Bonaguidi et al., 2012).

Adult-born DGCs are especially vulnerable to epileptogenic insults. Cells born in the weeks before and after an insult develop morphological and functional abnormalities (Parent et al., 2006; Jessberger et al., 2007; Walter et al., 2007; Murphy et al., 2011; Santos et al., 2011). Following epileptogenic insults, adult-born DGCs populate the dentate hilus (hilar ectopic granule cells), a region they rarely occupy in normal animals (Scharfman et al., 2000). Afferent inputs to these ectopic DGCs are abnormal, and the cells can exhibit atypical bursting properties (Zhan et al., 2010; Myers et al., 2013; Althaus et al., 2015). DGCs with basal dendrites are also common in the epileptic brain, a feature typically absent from nonepileptic rodent DGCs. Basal dendrites are hypothesized to form recurrent circuits and promote hyperexcitability within the hippocampus (Ribak et al., 2000; Shapiro et al., 2008).

Although it is well established that abnormal DGCs are derived from adult progenitor cells, it is not known whether all progenitors contribute equally to the production of abnormal cells, or whether distinct subsets of progenitors preferentially produce them. Answering this question will provide novel insights into the mechanisms underlying aberrant granule cell accumulation. Equal participation among progenitors suggests systemic changes in the factors regulating granule cell development, while unequal participation suggests regional disruption of neurogenic niches or intrinsic changes within individual progenitors. Here, we used a conditional Brainbow reporter line driven by an inducible $Gli1-CreER^{T2}$ promotor construct to trace the lineage of clones arising from Gli1-expressing granule cell progenitors in the pilocarpine model of epilepsy. Brains were rendered translucent using a novel clearing agent. Hippocampi were imaged in their entirety to identify and characterize groups of daughter cells, known as "clonal clusters," each of which originates from a single labeled progenitor.

## Materials and Methods

### Animals

All methods used involving animals were approved by the Institutional Animal Care and Use Committee of the Cincinnati Children's Hospital Research Foundation and conform to National Institutes of Health guidelines for the care and use of animals. For the present study, hemizygous $Gli1-CreER^{T2}$ mice (Ahn and Joyner, 2005; https://www.jax.org/strain/007913) were crossed to mice homozygous for a $Gt(ROSA)26Sor^{tm1(CAG-Brainbow2.1)Cle}$/J "Brainbow" reporter construct (Cai et al., 2013; https://www.jax.org/strain/013731) to generate double-transgenic $Gli1-CreER^{T2}::Brainbow$ mice. All animals were on a C57BL/6 background. A total of 30 double-transgenic mice were randomly assigned to the control or treatment [pilocarpine-induced status epilepticus (SE)] group for the present study. Postnatal tamoxifen treatment of these mice restricts $CreER^{T2}$ expression to type 1 cells in the hippocampal subgranular zone (Ahn and Joyner, 2005; Murphy et al., 2011; Hester and Danzer, 2013). Tamoxifen-induced activation of Cre-recombinase causes random excision and/or inversion between multiple pairs of lox sites, leading to the expression of one of four possible different fluorescent proteins in progenitor cells and all their progeny (Livet et al., 2007). To facilitate morphological analyses, only cells expressing the cytoplasmic red fluorescent protein (RFP) or yellow fluorescent protein (YFP) were examined. Cells expressing cyan fluorescent protein (CFP) were excluded because morphological details were poorly revealed by this membrane-bound protein. GFP-expressing cells were not observed in any of the animals, consistent with prior work (Calzolari et al., 2015).

### Tamoxifen-induced cell labeling and pilocarpine treatment

To achieve sparse labeling of progenitor cells, mice were given three injections of tamoxifen (250 mg/kg, s.c.) on alternate days during postnatal week 7 (Fig. 1A). At 8 weeks of age, all mice received methyl scopolamine nitrate in sterile saline solution (1 mg/kg, s.c.), followed 15 min later by either pilocarpine (420 mg/kg, s.c.; $n = 25$) or saline solution (controls, $n = 5$). Animals were monitored behaviorally for seizures and the onset of SE (defined as

Received August 12, 2015; accepted December 1, 2015; First published December 24, 2015.
¹The authors declare no competing financial interests.
²Author Contributions: S.P.S., C.L.L., and A.A. performed research; S.P.S., C.L.L., J.J.M., and S.C.D. analyzed data; S.P.S., C.L.L., J.J.M., and S.C.D. wrote the paper; S.C.D. designed research.
³This research was supported by National Institute of Neurological Disorders and Stroke Grant 2R01-NS-065020 (to S.C.D.), an Albert J. Ryan Foundation Fellowship (to S.P.S.), and an American Heart Association Pre-Doctoral Award (to S.P.S.).

Acknowledgments: The content is solely the responsibility of the authors and does not necessarily represent the official views of the National Institute of Neurological Disorders and Stroke, the National Institutes of Health, the Albert J. Ryan Foundation, or the American Heart Association. We thank Matthew Kofron and Micheal Muntifering (Confocal Imaging Core, Cincinnati Children's Hospital, Cincinnati, OH) for technical advice on imaging modalities used for this study. We also thank Keri Kaeding and Mary Dusing for useful comments on the earlier versions of this manuscript.

Correspondence should be addressed to Steve C. Danzer, 3333 Burnet Avenue, ML 2001, Cincinnati, OH 45229-3039. E-mail: steve.danzer@cchmc.org.
DOI:http://dx.doi.org/10.1523/ENEURO.0087-15.2015
Copyright © 2015 Singh et al.

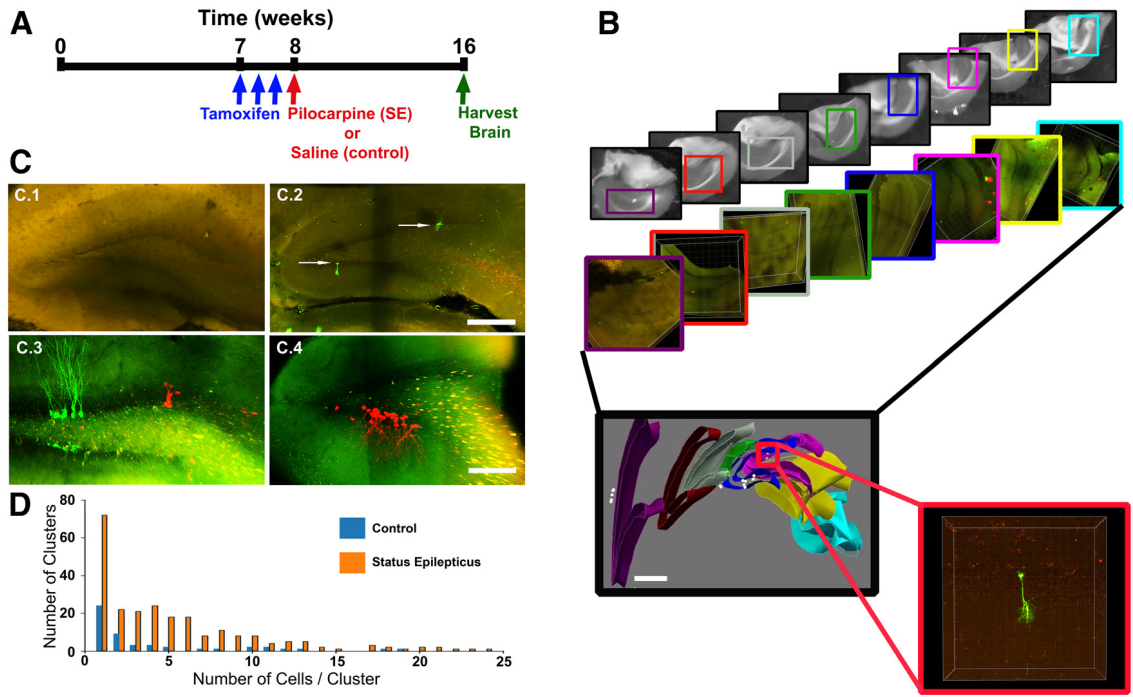

**Figure 1.** ***A***, Timeline depicting the experimental paradigm used. To induce fluorophore expression, mice were injected with tamoxifen three times during postnatal week 7 and subsequently underwent treatment with either pilocarpine or saline solution on postnatal week 8. Mice were killed on postnatal week 16. ***B***, Example of a three-dimensional reconstruction of the mouse hippocampus. The Scale-cleared 300 $\mu$m sections were imaged, aligned, and reconstructed into a three-dimensional reconstruction of the hippocampus with single-cell resolution. ***C.1***, Brainbow fluorophore expression was absent from animals not treated with tamoxifen. ***C.2***, A small cohort of animals was killed 2 d after the last tamoxifen injection (in week 7), and analysis of their dentate gyri revealed that the tamoxifen treatment induced, on average, two type 1 cells (indicated by white arrows) per 300 $\mu$m hippocampal section. ***C.3***, ***C.4***, Clonal clusters were observed in both control (***C.3***) and pilocarpine-treated (***C.4***) animals. ***D***, The number of cells per cluster increased in pilocarpine-treated SE animals. Scale bars: ***B*** (three-dimensional reconstruction), 600 $\mu$m; ***C.1***, ***C.2***, 250 $\mu$m; ***C.3***, ***C.4***, 200 $\mu$m.

continuous tonic–clonic seizures). Following 3 h of SE, mice were given two injections of diazepam, 10 min apart (10 mg/kg, s.c.), to alleviate seizure activity. Mice were given sterile Ringer's solution as needed to restore pretreatment body weight and were then returned to their home cages, where they were provided with food and water *ad libitum*. Mice were housed under a 14 h/10 h light/dark cycle to optimize breeding (Fox et al., 2007). Of the 25 mice randomly assigned to pilocarpine treatment, 12 (48%) survived to the end of the experiment, yielding a final group consisting of 7 males and 5 females. There was no mortality among the five control animals (three males and two females).

**Tissue preparation for whole hippocampal imaging**
At 16 weeks of age, animals were killed by intraperitoneal injection of 100 mg/kg pentobarbital. The mice were perfused through the ascending aorta with ice-cold PBS (0.1 M PBS) containing 1 U/ml heparin for 30 s at 10 ml/min, immediately followed by a 2.5% paraformaldehyde plus 4% sucrose solution in 0.1 M PBS at 25°C for 10 min. Brains were removed, dissected into left and right hemispheres, and post-fixed in the same solution overnight at 4°C. Brain hemispheres were cryoprotected in 10%, 20%, and 30% sucrose in PBS for 24, 24, and 48 h, respectively. The hemispheres were then frozen in isopentane cooled to −23°C with dry ice and stored at −80°C until sectioning. Brain hemispheres were thawed in PBS for scale clearing, and 300 $\mu$m coronal sections were cut on a tissue slicer (Campden Instruments). Sections were transferred to 24 multiwell tissue culture plates (Becton Dickinson), maintaining their septotemporal order. Sections were incubated for optical clearing in ScaleA2 for 2 weeks at 4°C (Hama et al., 2011). At least one hemisphere from each animal was used for clonal analysis, and in three cases both hemispheres were used (2 SE and 1 control). No significant differences in cluster composition were found between hemispheres within animals [Mann–Whitney rank sum test (RST)], so hemispheres were pooled by animal for statistical analysis. The remaining hemispheres were used for immunohistochemical characterization of Brainbow-labeled cells.

**Tissue preparation for immunohistochemistry**
Unused hemispheres from a subset of animals ($n = 5$) were sectioned coronally on a cryostat at 60 $\mu$m and mounted to gelatin-coated slides. Sections were immunostained with mouse anti-nestin (1:100; Millipore), chicken anti-glial fibrillary acidic protein (GFAP; 1:500; Chemicon), goat anti-doublecortin (1:250; Santa Cruz Biotechnology), mouse anti-calretinin (1:200; Millipore) or guinea pig anti-calbindin-D-28K (1:200; Sigma-Aldrich).

Alexa Fluor 405 goat anti-mouse, 488 goat anti-chicken, 594 goat anti-mouse, 647 donkey anti-goat or Alexa Fluor 647 goat anti-guinea pig secondary antibodies were used (Invitrogen). Tissue was dehydrated in alcohol series and cleared in xylenes, and coverslips were secured with mounting media (Krystalon, Harleco).

### Confocal microscopy

ScaleA2-cleared hippocampal sections were imaged on an A1R GasAsP confocal system attached to a motorized Eclipse Ti inverted microscope (Nikon Instruments). This system was used to capture three-dimensional image stacks through the z-depth of the tissue at 1 $\mu$m steps using a 10× Plan Apo λ objective (NA, 0.25) at 1× optical zoom (field size 1024 × 1024 pixels, 1.23 pixels/$\mu$m). These 10× image stacks were used to identify clonal clusters, defined here as cells expressing the same fluorophore and contained within a 150 $\mu$m radius of the clone center (Bonaguidi et al., 2011; Calzolari et al., 2015). Identified clonal clusters were then imaged using a 40× Plan Apo infrared- differential interference contrast water immersion objective (NA, 1.3) at 1× optical zoom (field size, 1024 × 1024 pixels, 0.31 pixels/$\mu$m). All cells selected for analysis were brightly labeled with RFP or YFP and had their somas fully contained within the tissue section. The investigator was blind to treatment group during all image collection and data analysis.

### Three-dimensional hippocampal reconstructions

Confocal z-series image stacks were converted into 8 bit RGB.tiff files. Reconstruct Software (John C. Fiala, the National Institutes of Health; Lu et al., 2009) was used to septotemporally align sections (10× images) for each hippocampus (Fig. 1B). Aligned z-stacks were imported into Neurolucida software for analysis (version 11.01, Microbrightfield). Borders of the granule cell body layer were traced at z-intervals of 100 $\mu$m to recreate the whole hippocampus.

### Morphological classification

Higher-magnification images (40×) were used to categorize cells within each cluster as follows: (1) type 1 cell, with a small cell body located in the subgranular zone and a single, radial process that projects through the granule cell layer and terminates in the inner molecular layer (type 1 progenitor cells express nestin and GFAP); (2) type 2/3 cells, with a cell body located in the subgranular zone and short, aspiny processes projecting parallel to the plane of the granule cell body layer (because their appearance is morphologically similar, we did not attempt to distinguish between type 2 and 3 cells, or the different subtypes of type 2 cells; type 2 and 3 cells express the cellular proliferation marker doublecortin); (3) immature granule cells, with a cell body in the granule cell body layer and aspiny dendrites that project radially through the granule cell body layer, but terminate prior to reaching the hippocampal fissure (typically with growth cones at the tips; these cells occasionally possessed short, aspiny basal dendrites and express calretinin); (4) normal mature granule cells, with their somas located in the granule cell body layer and spine-coated dendrites projecting to the hippocampal fissure (mature granule cells express calbindin); (5) hilar ectopic granule cells, with spiny dendrites and a cell body located in the hilus (at least two cell body distances, ~20 $\mu$m, away from the granule cell layer-hilar border); (6) mature granule cells with basal dendrites, possessing all the features of normal mature granule cells (see point 4), but with at least one dendrite originating from the hilar side of the soma [i.e., arising from a region below the soma midline; cells with basal dendrites projecting into either the dentate hilus, or the dentate molecular layer (recurrent basal dendrites), were included in this measure. Only granule cells with clearly visible axons were scored for basal dendrites. Basal dendrites are often thin and difficult to visualize in deeper regions of the tissue. Well developed axons should be present on all mature granule cells. Confirming that these axons can be visualized limits the entry of false negatives into the dataset. If the axon can be visualized, then any basal dendrites, which are typically of higher caliber than the axon, should also be detectable]; and (7) astrocytes, defined as cells with a small cell body, located anywhere within the dentate gyrus and possessing numerous thin, aspiny process projecting outward in a stellate fashion (for review of granule cell developmental markers, see Kempermann et al., 2004, 2015; Bonaguidi et al., 2012).

### Statistical analysis

A Microsoft SQL Server (version 2012) was used to query the dataset for different clone compositions, and statistical analysis was performed using R (version 0.98.109) or SigmaPlot (version 12.5). Sex and treatment effects were determined using two-way ANOVA. Individual group differences were determined using the Holm–Sidak method for all ANOVA results. Parametric tests were used for data that met assumptions for normality and equal variance. Data that failed assumptions of normality and equal variance were either transformed, as noted in the text to meet these assumptions, or were analyzed using nonparametric alternative tests. The actual tests used are noted in the text. Values presented are the mean ± SEM (least square means for two-way ANOVA data) or median (range), as appropriate. Details of statistical tests are given in Table 1.

The statistical analysis for the frequency/distribution of ectopic cells and basal dendrites was performed using the binomial distribution (to compute probabilities of combinatorial events). The experiment-wise error was conservatively set at 0.001 (Cumming, 2010). Corrections for multiple comparisons were performed using a Bonferroni correction. For clusters containing ectopic cells, the resultant $p$ value for significance for the pilocarpine-treated animals was calculated to be $4.17 \times 10^{-6}$ (0.001/240). Similarly, for clusters containing DGCs with basal dendrites, the probability of a single trial success was 0.0614, and the critical $p$ value was calculated to be $5.41 \times 10^{-6}$ (0.001/185).

### Figure preparation

Maximum projections from z-series stacks were prepared using NIS-Elements Ar Microscope Imaging Software (version 4.0). Contrast, brightness, montage adjustments, and figure preparation were performed using Adobe Photoshop CS5 (version 12.0). Identical filtering and adjustments to brightness and contrast were performed for images meant for comparison. Tableau (version 8.0) and Microsoft Excel (version 2013) were used to create graphs and visualizations. The image in Figure 4 was cropped to remove neuronal structures above and below the cell of interest that would otherwise obscure it (Walter et al., 2007; McAuliffe et al., 2011; Murphy et al., 2012). This image is best viewed as a neuronal reconstruction, similar to traditional neuroanatomical techniques (Danzer et al., 1998), rather than a standard photomicrograph.

**Table 1: Statistical tests**

| Data | Data structure | Type of test | Power |
|---|---|---|---|
| Number of clonal clusters per hippocampus | Normality test (Shapiro-Wilk): passed ($p = 0.561$)<br>Equal variance test (Brown-Forsythe): passed ($p = 0.614$) | Two-way ANOVA with treatment and sex as factors | Treatment: 0.372<br>Treatment × sex: 0.579 |
| Mean clone size | Normality test (Shapiro-Wilk): passed ($p = 0.658$)<br>Equal variance test (Brown-Forsythe): passed ($p = 0.399$) | Two-way ANOVA with treatment and sex as factors | Treatment: 0.516<br>Treatment × sex: 0.259 |
| Control vs SE, type 1 | Normality test (Shapiro-Wilk): failed ($p < 0.050$) | Mann–Whitney RST | 25–75% CIs, control: 7.76–37; SE: 0–3.5 |
| Control vs SE, type 2 | Normality test (Shapiro-Wilk): failed ($p < 0.050$) | Mann–Whitney RST | 25–75%, control: 0–21.6; SE: 0–6.1 |
| Control vs SE, immature cells | Normality test (Shapiro-Wilk): failed ($p < 0.050$) | Mann–Whitney RST | 25–75%, control: 0–23.4; SE: 0–4.5 |
| Control vs SE, mature cells | Normality test (Shapiro-Wilk): failed ($p < 0.050$) | Mann–Whitney RST | 25–75%, control: 26.9–83.6; SE: 75.0–93.8 |
| Control vs SE, astrocytes | Normality test (Shapiro-Wilk): failed ($p < 0.050$) | Mann–Whitney RST | 25–75%, control: 0–10.4; SE: 3.2–7.0 |
| Clusters with progenitors | Normality test (Shapiro-Wilk): passed ($p = 0.526$) | $t$ test | Power = 0.793 |
| Self-renewing (2 type 1) clusters | Normality test (Shapiro-Wilk): failed ($p < 0.050$) | Mann–Whitney RST | 25–75%, control: 0–0.15; SE: 0–0 |
| Fully differentiated clusters | Normality test (Shapiro-Wilk): passed ($p = 0.536$) | $t$ test | Power = 0.790 |
| Ectopic cells | Normality test (Shapiro-Wilk): failed ($p < 0.050$) | Mann–Whitney RST | 25–75%, control: 0–0; SE: 0–0 |
| Cells with basal dendrites | Normality test (Shapiro-Wilk): failed ($p < 0.050$) | Mann–Whitney RST | 25–75%, control: 0–0; SE: 0–0 |

## Results

### *In vivo* lineage tracing of individual Gli1-expressing progenitor cells in the adult mouse hippocampus

To study the proliferative activity of a cohort of Gli1-expressing granule cell progenitors in epilepsy, we treated double-transgenic *Gli1-CreER$^{T2}$::Brainbow* reporter mice with tamoxifen at postnatal week 7 to trace the lineage of these cells. A small cohort of animals was perfused 2 d later, revealing an average of two type 1 cells per 300 μm hippocampal coronal section (Fig. 1), an optimal labeling sparsity for identifying individual clones (Bonaguidi et al., 2011). Gli1 expression has been shown to mark multipotent type 1 stem cells, which give rise to type 2/3 stem cells and other differentiated progeny (Encinas et al., 2011). Animals in the main study groups received either saline or pilocarpine 1 week after tamoxifen treatment (Fig. 1A). Pilocarpine induces acute status epilepticus (SE) and the subsequent development of epilepsy a few weeks later (Turski et al., 1983). Animals were killed 2 months after pilocarpine treatment, when spontaneous seizures are typically frequent (Castro et al., 2012; Hester and Danzer, 2013). Hippocampi were imaged in their entirety to identify all fluorescently labeled cells (Fig. 1B). Brainbow fluorophore expression was strictly tamoxifen dependent; and was limited to dentate granule cells, astrocytes, and sub-granular zone progenitor cells (Fig. 1C). We first assessed whether SE altered the number of clonal clusters per hippocampus (control, female, $n = 2$ mice, 4.0 ± 5.8 clusters per hippocampus; control, male, $n = 3$, 13.3 ± 4.8; SE, female, $n = 5$, 24.9 ± 4.2; SE, male, $n = 7$, 11.0 ± 2.9). The effect of SE was found to be dependent on animal sex (Fig. 2; $p = 0.025$, two-way ANOVA). *Post hoc* tests revealed a significant increase in female, but not male, mice in clusters per hippocampus relative to controls ($p = 0.012$, Holm–Sidak method) and significantly more clusters in females versus males within the pilocarpine-treated groups ($p = 0.017$, Holm–Sidak method). Differences between sexes could reflect differential numbers of progenitors prior to pilocarpine treatment and differential behavior of progenitors after treatment. Greater apoptosis of quiescent progenitor cells or entire clonal groups in males, for example, would reduce the number of clonal clusters. Sexually dimorphic changes in neurogenesis have been observed following early-life stress in rodents (Loi et al., 2014). The present findings suggest that dimorphic responses to adult SE also occur.

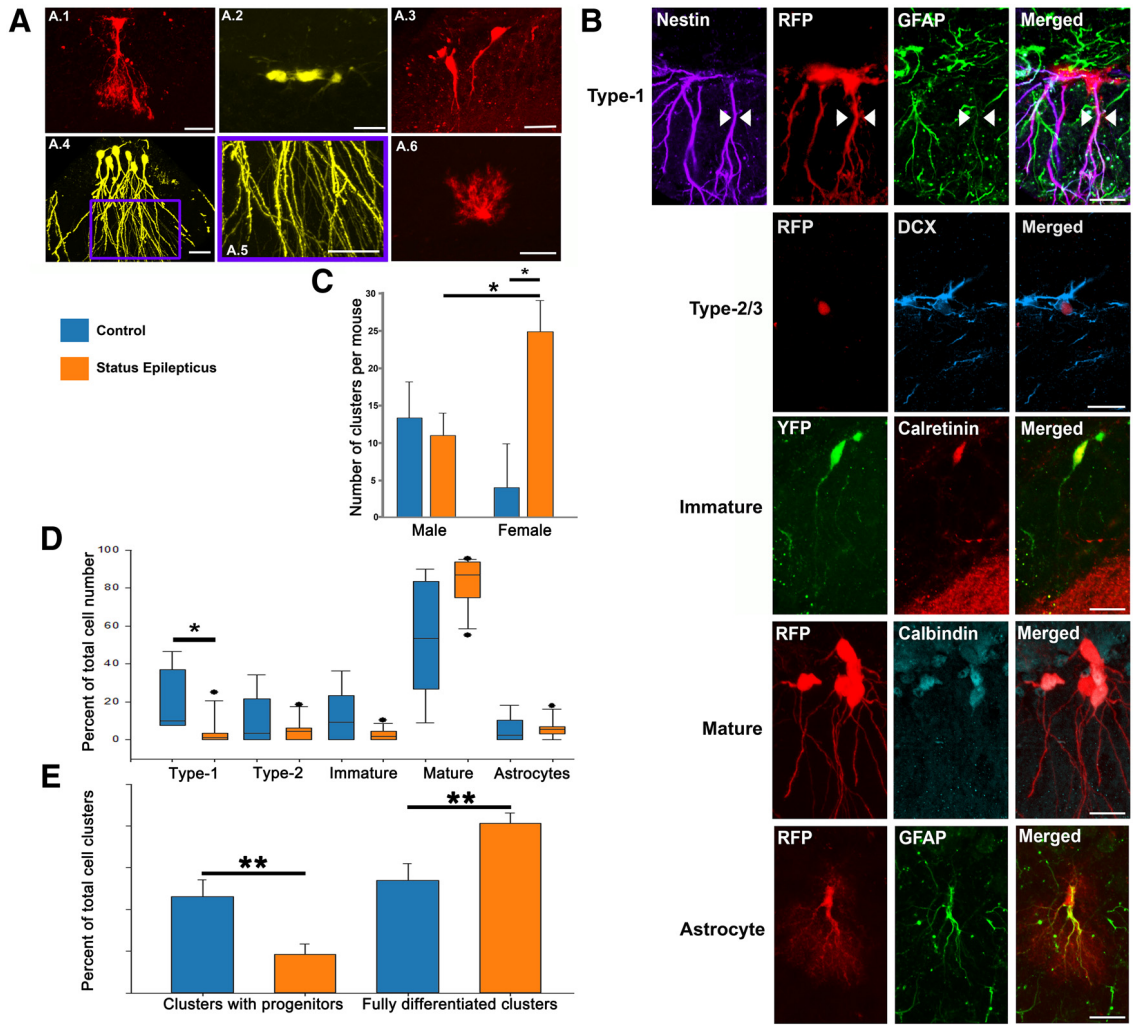

**Figure 2.** *A*, Cells present in clonal clusters were classified based on morphology (see Materials and Methods) into type 1 progenitor cells (*A.1*); type 2/3 progenitor cells (*A.2*); immature granule cells (*A.3*); mature granule cells (*A.4*), with spiny apical dendrites (*A.5*); or astrocytes (*A.6*). *B*, Immunocharacterization of the different cell types. Type 1 cells were shown to express nestin and GFAP; type 2/3 cells expressed doublecortin (DCX); immature DGCs expressed calretinin; mature DGCs expressed calbindin; and astrocytes were shown to express GFAP. *C*, The number of clonal clusters per mouse hippocampus was significantly increased in female mice that underwent SE relative to female controls. Female SE mice also had more clusters than male SE mice. *D*, Graph shows the composition of cell types in clonal clusters from control and SE animals. There was a significant decrease in the number of type 1 cells and a trend ($p = 0.06$) toward an increase in the number of mature cells in mice exposed to status. *E*, The percentage of clusters containing either type 1 or type 2/3 progenitors was decreased in SE mice relative to controls, while the percentage of fully differentiated clusters increased. $*p < 0.05$; $**p < 0.01$. Scale bars: *A.1–A.3*, *A.6*, 25 μm; *A.4*, *A.5*, 50 μm; *B*, 20 μm.

## Status epilepticus increases the average number of cells per clonal cluster

Increased hippocampal neurogenesis and cell survival have been consistently demonstrated in epilepsy models (Bengzon et al., 1997; Parent et al., 1997; Gray and Sundstrom, 1998; Parent et al., 1998). In addition to an increase in the number of clonal clusters in female mice, the present work also revealed an increase in the mean size of individual clones. Specifically, mean clonal size increased from $3.0 \pm 0.7$ cells/cluster in controls to $5.1 \pm 0.5$ in SE mice [Fig. 1, $p = 0.033$, SE mice ($n = 12$) vs control mice ($n = 5$), two-way ANOVA]. In contrast to cluster number, however, no differences between males and females were found for cluster size ($p = 0.822$), nor was there an interaction between treatment and sex ($p = 0.107$). Similarly, no additional sex differences or interactions were found in pretests for all additional data presented here (data not shown), so males and females were binned for all further statistical analyses. Together, these data suggest that the increase in new granule cells after SE is likely due to increased proliferation among individual progenitors (increased cluster size) and/or reduced apoptosis of their progeny (more clusters, increased cluster size).

## Status epilepticus promotes terminal differentiation of hippocampal progenitor cells

While neurogenesis is increased after an acute epileptogenic injury, it can decrease in chronic epilepsy (Hattian-gady et al., 2004; Danzer, 2012). A growing body of literature demonstrates that progenitor cell pools can be depleted as progenitors proceed through multiple rounds of division, ultimately leading to terminal differentiation (Ledergerber et al., 2006; Encinas et al., 2011). Reduced neurogenesis in chronic epilepsy, therefore, could be a direct consequence of increased progenitor cell activity early in the disease.

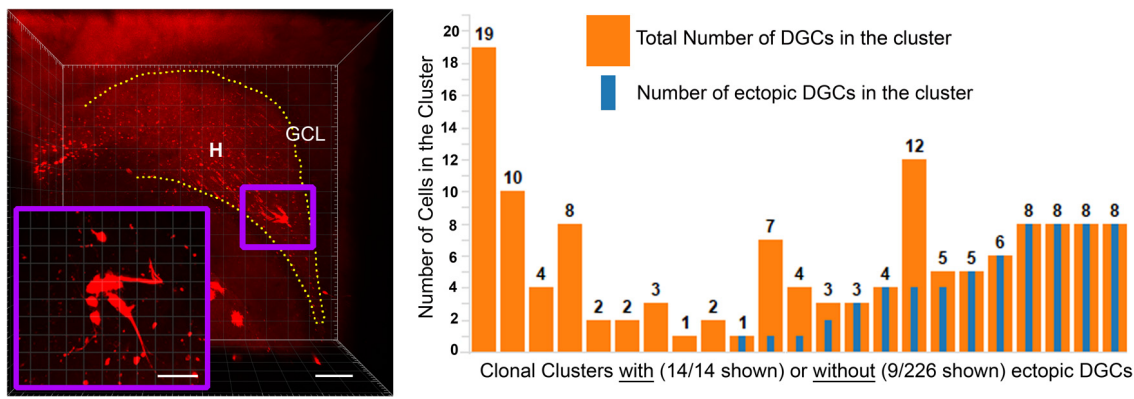

**Figure 3.** Ectopic dentate granule cells are derived from a small number of clonal clusters. Shown is an image of clonal cluster composed entirely of hilar ectopic dentate granule cells (higher-magnification image is outlined in purple in the inset). The graph shows quantification of all the clusters from SE animals that contained ectopic cells. Additionally, for comparison, nine randomly selected clusters containing no ectopic cells are shown. Orange bars show the total number of cells in the cluster, whereas the blue bars show the number of ectopic cells. Ectopic DGCs tended to occur in clusters in which majority of the cells are ectopic. GCL, granule cell layer; H, hilus. Scale bars: ***A***, 150 μm; ***A*** inset, 40 μm.

To assess whether this might be the case, we used morphological criteria to classify cells as type 1 progenitors, type 2/3 progenitors, immature granule cells, mature granule cells, or astrocytes (Fig. 2A). The accuracy of this morphological classification was confirmed by immunocharacterization of the different cell types quantified (Fig. 2B). This lineage analysis revealed a significant reduction in the progenitor cell pool in SE-exposed animals. Compared with control animals, SE animals exhibited an 84% reduction in type 1 cells [$p = 0.013$, SE mice ($n = 12$) vs control mice ($n = 5$), Mann–Whitney RST]. No difference in the percentage of type 2/3 cells was found [$p = 0.870$, SE mice ($n = 12$) vs control mice ($n = 5$), Mann–Whitney RST]. Overall, there was a 58% reduction in the percentage of clusters containing either type 1 or type 2/3 progenitors, dropping from 46.2 ± 8.1% in control animals to 18.6 ± 5.0% in animals exposed to SE [Fig. 2E; $p = 0.009$, SE mice ($n = 12$) vs control mice ($n = 5$), two-tailed $t$ test]. Symmetric self-renewing clusters (composed of two type 1 cells) decreased from a median of 4.5% (range, 0-16.7%) of clusters/animal in controls to nil in SE animals [$p = 0.006$, SE mice ($n = 12$) vs control ($n = 5$), Mann–Whitney RST]. The significant shift away from progenitors in animals exposed to SE was mirrored by a nonsignificant trend in the proportion of mature granule cells produced [$p = 0.065$, SE mice ($n = 12$) vs control mice ($n = 5$), Mann–Whitney RST] and a significant increase in the percentage of clusters that were "fully differentiated," meaning the clusters contained only granule cells and astrocytes, and were devoid of progenitors [$p = 0.010$, SE mice ($n = 12$) vs control mice ($n = 5$), two-tailed $t$ test). No differences in the proportions of immature granule cells ($p = 0.299$, Mann–Whitney RST) or astrocytes ($p = 0.223$, Mann–Whitney RST) were found.

### Ectopic DGCs appear in clonal clusters in which the majority of the cells are ectopic

Hilar ectopic granule cells are a common pathology seen in temporal lobe epilepsy. These neurons are newly generated, arising after an epileptogenic brain insult, and are implicated in the development of epilepsy (Hester and Danzer, 2013). The percentage of newborn cells found in the hilar region of SE-exposed animals was significantly increased relative to controls [Fig. 3; $p = 0.004$; SE mice, 63 of 1238 DGCs (5.09%), control mice, 1 of 192 cells (0.52%); Mann–Whitney RST], which is consistent with the findings of previous studies (Parent et al., 2006). Clonal analysis revealed that ectopic cells were concentrated within a small number of clusters. Specifically, the 63 identified ectopic cells from animals exposed to status were contained within only 14 clusters (Fig. 3B). Within these clusters, 76.8% of all cells were ectopic (Fig. 3B).

The probability of finding a set number of ectopic cells, $S$, in a cluster containing a total of $N$ cells was computed for all values of $S$ from 1 to $N$ (e.g., when $S = N$, 100% of the cells in the cluster are ectopic). The minimum number of ectopic cells $S$, at which the $p$ value reaches the target $p$ value for significance ($p = 4.17 \times 10^{-6}$; see Materials and Methods for calculations) was determined for every cluster and was compared with the observed number of ectopic cells. We found that 6 of the 14 clusters with ectopic cells exceeded the threshold $p$ value for significance (binomial, $p < 4.17 \times 10^{-6}$; Fig. 3). The binomial probability of finding 6 of 240 clusters exceeding this value is statistically minute ($<1.33 \times 10^{-21}$). Finding one cluster that exceeded the value would indicate that ectopic cells are not randomly distributed, and in the present study six such clusters were observed. These results provide overwhelming evidence that the accumulation of ectopic cells in certain clusters is not a random event.

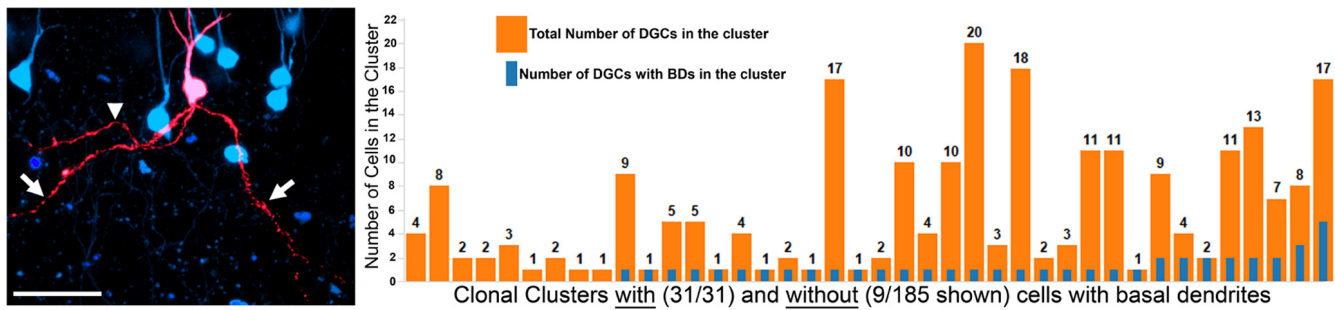

**Figure 4.** Dentate granule cells with basal dendrites arise from diverse clonal clusters. Shown is a neuronal reconstruction of a granule cell (red) with basal dendrites (white arrows) within a clonal cluster. The axon is denoted by the arrowhead. Adjacent cells in the cluster are shown in blue. The graph shows quantification of all the clusters from SE animals that contained cells with basal dendrites (blue bars) relative to total cluster size (orange bars). For comparison, a subset of randomly selected clusters that contained only normal DGCs is shown. Scale bar, 50 $\mu$m.

**DGCs with basal dendrites occur in clonal clusters in which the majority of the cells do not have a basal dendrite**

Another common pathology observed in the epileptic brain is the presence of dentate granule cells with basal dendrites (Spigelman et al., 1998; Buckmaster and Dudek, 1999). In the current study, 6.14% of granule cells from SE-exposed animals possessed basal dendrites. By contrast, only one cell with basal dendrites was found among the control animals ($p = 0.034$; SE mice, 43 of 700 DGCs; control mice, 1 of 101 cells; Mann–Whitney RST). The 43 basal dendrite-possessing granule cells from SE mice were distributed among 31 of 185 clonal clusters (Fig. 4). Among these 31 clusters that contained a cell with a basal dendrite, 23 had only a single basal dendrite-possessing cell, 6 had two cells with basal dendrites, 1 had three cells with basal dendrites, and 1 had five cells with basal dendrites (Fig. 4). Using the significance criterion described in Materials and Methods (similar to that used for ectopic cells), there were no clusters that contained a significant number of DGCs with basal dendrites (Fig. 4). Indeed, even the biggest cluster—containing 17 DGCs with 5 harboring a basal dendrite—failed to reach significance even if the experiment-wise $\alpha$ value is relaxed to 0.05 ($p = 0.054$; threshold $p$ value for $\alpha = 0.001$, $5.41 \times 10^{-6}$; for $\alpha = 0.05$, $2.60 \times 10^{-4}$).

**Clonal clusters have more type 1 cells in dorsal hippocampus**

Previous studies (Jinno, 2011; Kheirbek and Hen, 2011; Jhaveri et al., 2015) have reported topographical differences within the hippocampus with respect to neurogenesis, cell densities, functional properties, and electrophysiological properties. To determine whether there are dorsal–ventral differences in progenitor cell behavior, we correlated cluster composition with cluster bregma level (Paxinos and Franklin, 2001). Within SE animals, the number of type 1 cells was significantly correlated with bregma level, with greater numbers of cells identified in more dorsal regions (Fig. 5; $p = 0.0037$, Spearman rank order correlation). This effect persisted when bregma levels were correlated with the proportion of type 1 cells at each level (type 1 cells/total cells), suggesting that differences in total cell numbers cannot account for the finding ($p = 0.0238$, Spearman rank order correlation). The numbers and proportions of mature, immature, type 1, type 2, astrocytic, ectopic, and basal dendrite-possessing cells were not significantly correlated with bregma level (data not shown). No significant correlations were found between bregma level and the number or proportion of any cell types in control animals (data not shown).

## Discussion

Abnormal hippocampal granule cells are common in animal models of temporal lobe epilepsy (Rolando and Taylor, 2014) and in tissue from patients with the disease (Sutula et al., 1988; Parent et al., 2006). Prior studies (Walter et al., 2007; Kron et al., 2010) have established that many of these abnormal cells are adult generated. In the present study, we queried whether two important abnormalities, ectopic DGCs and DGCs with basal dendrites, are derived with equal likelihood from the entire progenitor pool, or whether they are preferentially produced by a subset of progenitors. We found that ectopic granule cells were highly concentrated within distinct clonal clusters, with many containing only ectopic cells. By contrast, cells with basal dendrites were relatively evenly distributed among clones. These findings strongly suggest the existence of distinct mechanisms regulating ectopic cell migration and basal dendrite formation. A second key finding provides new insights into the bimodal changes in neurogenesis rates observed in epileptic animals. Neurogenesis increases acutely following an epileptogenic insult; however, animals with chronic epilepsy exhibit reduced neurogenesis. Depletion of the progenitor pool—potentially as a direct consequence of the early increase in neurogenesis—has been hypothesized to account for these changes (Hattiangady et al., 2004). Our data provide direct evidence that this is occurring, with a 70% increase in the number of daughter cells/clone, and a corresponding decrease in the percentage of actively dividing and self-renewing clones (Fig. 6).

**Limitations of the current study**

The present study uses a clonal analysis methodology previously validated for the dentate gyrus (Bonaguidi et al., 2011). Nonetheless, it is expected that progenitor cells labeled with the same fluorophore will occasionally appear in close proximity to one another, leading to the false conclusion that they represent a single clonal cluster. Conclusions based on rare clonal events, therefore, should be made with caution if the observation could also be accounted for by a false merging of clusters. As an example, it is possible that some of the clones with mixes of ectopic and normally positioned cells shown in Figure 3 are actually merged clonal clusters. Notably, however, our findings in control animals are remarkably similar to those of Bonaguidi et al. (2011) using the Nestin-CreER$^{T2}$ driver line to label progenitor cells. We found a similar distribution of cluster sizes (Fig. 1D) in the animals and were able to reproduce key findings, such as the occurrence of symmetric cell divisions, yielding two type 1 cells. One difference we noted was a greater degree of neurogenesis among clusters using the Gli1 driver relative to the nestin driver used previously. The nestin driver produced a roughly equal ratio of neurons to astrocytes (Bonaguidi et al., 2011; Song et al., 2012); whereas, neurons were much more common in the present work (Fig. 2D). The higher ratio of neurons to astrocytes is consistent with studies using cell birth-dating and virus-labeling approaches (Steiner et al., 2004), and may reflect differences between Gli1- and nestin-expressing stem cells.

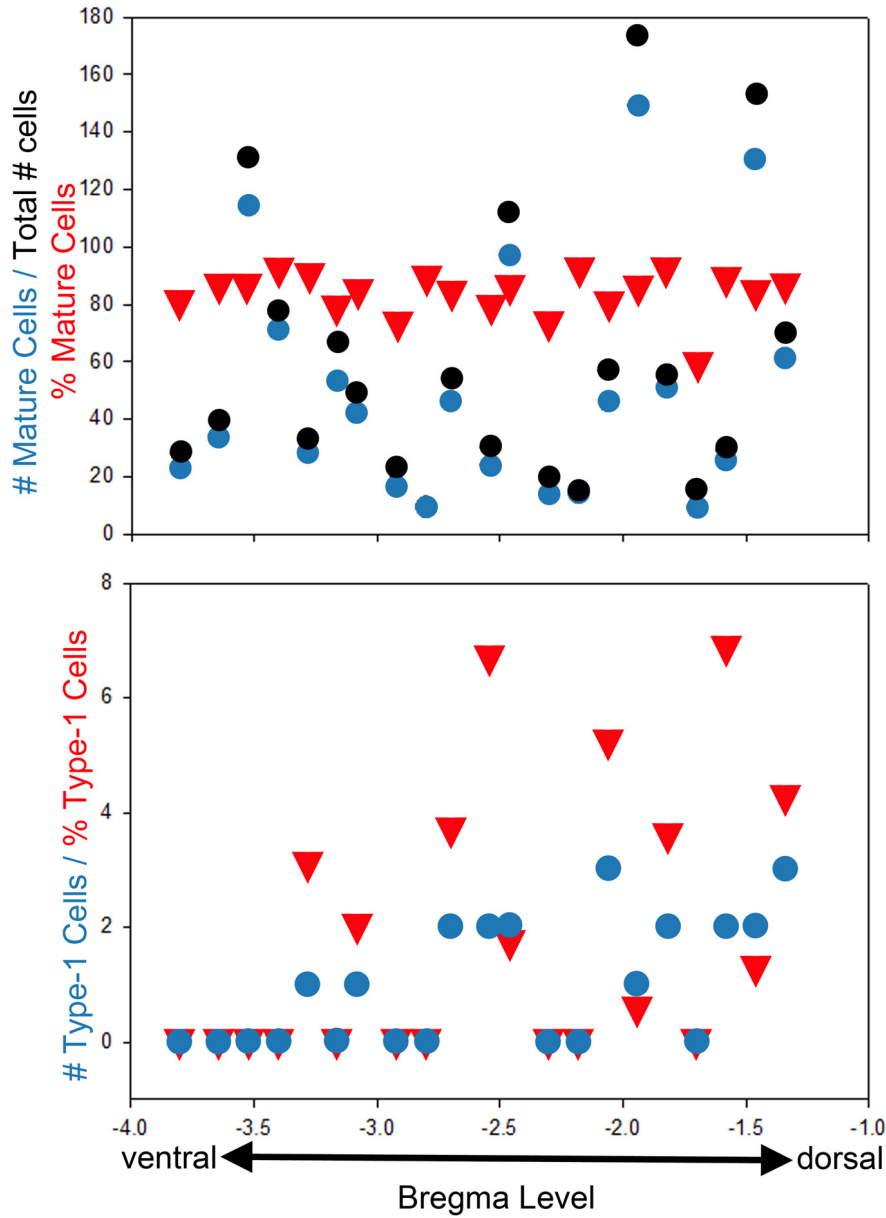

**Figure 5.** Graphs show the distribution of mature granule cells (top) and type 1 cells (bottom) along the dorsal–ventral axis of the hippocampus. Black dots depict the total number of cells at each bregma level (top graph only), whereas blue dots depict the number of mature or type 1 cells, respectively. Red triangles give the percentage of mature or type 1 cells at each level. No relationship between mature cells and bregma level was evident, while higher numbers and proportions of type 1 cells were present at more dorsal levels.

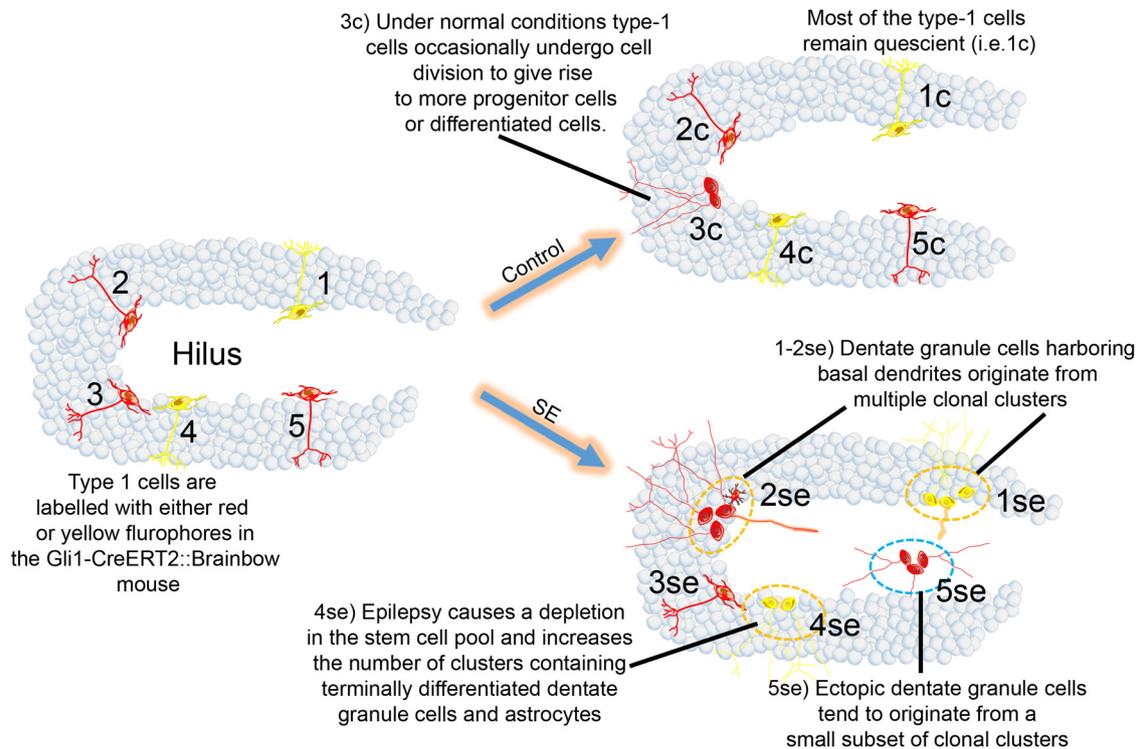

**Figure 6.** Summary of the key findings of the study. The first panel shows five type 1 progenitor cells labeled with either RFP (red) or YFP (yellow), numbered from 1 to 5. Under control conditions, most of the type 1 cells remain quiescent (progenitor cells 1, 2, 4, and 5 remain quiescent); however, a proportion of type 1 cells will enter the mitotic cycle to give rise to differentiated cells (only progenitor cell 3 undergoes terminal differentiation). After epileptogenesis, the following three key changes occur: (1) the number of clusters containing type 1 cells decreases in epileptic animals relative to controls, and the number of clusters composed of differentiated DGCs and astrocytes increases (progenitors 1, 2, 4, and 5 terminally differentiate); (2) progenitor cells either produce all correctly located offspring, or ectopic offspring (progenitor cell 5 gives rise to a cluster composed entirely of ectopic DGCs); and (3) progenitor cells that produce correctly located offspring occasionally produce cells with a basal dendrite, but mostly produce cells with normal dendrites (progenitor cells 1 and 2 give rise to clusters that contain DGCs with basal dendrites and normal DGCs).

A second notable caveat is that the present study examined only progenitor cells labeled in the week before SE, in order to examine the impact of status epilepticus on progenitor cells (rather than their offspring, as would be achieved with earlier labeling), and to ensure equivalent labeling of progenitor populations between control and epileptic animals. The progenitor pool changes after SE, so labeling after the insult will presumably label a population of progenitors that diverges from controls. Aberrant granule cell integration occurs over a protracted time course, and includes immature cells born weeks before SE, as well as cells born months later. Whether the current findings will extend to populations generated at other time points remains to be determined.

**Clonal analysis of adult subgranular zone neurogenesis following status epilepticus**

In the present study, we used an *in vivo*, genetic, sparse-labeling approach to mark stem cells for lineage tracing. This approach has been used previously to study neural stem cell behavior in the subventricular (Calzolari et al., 2015) and subgranular (Suh et al., 2007; Bonaguidi et al., 2011) proliferative zones in healthy animals. We combined this approach with recently developed tissue-clearing protocols, allowing us to generate three-dimensional reconstructions of the entire rodent hippocampus; and the Brainbow reporter line, allowing us to separate clonal groups by fluorochrome expression. Poor recombination with the Brainbow reporter in CNS tissue limited our study to two colors, rather than the potential seven colors evident in other tissues (Livet et al., 2007; Cai et al., 2013). Even with two colors, however, the strategy provided sufficient spatial resolution for the study.

An additional advantage of the genetic approach is that it avoids disturbing the target tissue with direct brain injections, as is needed for retroviral strategies (Hope and Bhatia, 2011; Ming and Song, 2011). Retrovirus also targets $Sox2^+$ type 2 progenitor cells (Suh et al., 2007), while the Gli1 driver used here targets the parent type 1 cells (Ahn and Joyner, 2005), so the different strategies provide complimentary data.

**Localized regulation of ectopic granule cell formation**

Ectopic granule cells have been observed in a number of different epilepsy models. They are hyperexcitable (Scharfman et al., 2000; Althaus et al., 2015) and have atypical connections within the hippocampal network (Scharfman and Pierce, 2012), and their numbers correlate with the severity and duration of seizures (Hester and Danzer, 2013). The mechanisms underlying ectopic cell migration, however, are unknown. A mechanistic understanding would provide new insights into the development of therapeutic strategies for epilepsy. A putative mechanism that could account for the current results is mislocation of progenitor cells from the subgranule zone to the hilus during epileptogenesis (Parent et al., 2006). Alternatively, epileptogenic stimuli could activate "dormant" progenitors trapped in the hilus during development (Gaarskjaer and Laurberg, 1983; Scharfman et al., 2007). If one further presumes that daughter cells produced by hilar progenitors would not have access to the necessary cues directing them to migrate into the granule cell layer, the presence of entirely ectopic clonal groups could be accounted for. Alternatively, seizures might lead to the localized disruption of migratory cues, like reelin (Teixeira et al., 2012). Progenitor cells active in regions with disrupted cues would produce daughter cells that fail to migrate correctly, while progenitors in regions with intact cues would produce normal offspring. In support of this possibility, Parent et al. (2006) observed trains of cells migrating on glial scaffolds into the hilus after seizures, suggesting that localized changes in non-neuronal cells might play a role. Additional studies will be needed to distinguish among these possibilities.

The finding that clonal groups with ectopic cells tend to be made up of entirely ectopic cells is consistent with a Markov chain model (Lange, 2003). In a Markov model, there are two states a progenitor cell can assume: progenitors in state 1 give rise to DGCs correctly located in the cell body layer; whereas progenitors in state 2 give rise to ectopic DGCs. For the model, we assumed that, at every mitotic cycle, cells could either stay in the same state or transition between states. The transition matrix specifies the probabilities of these transitions. Our data show that the probability of a progenitor switching states is very low, and the probability that a progenitor will remain in the same state is close to 100%, implying that cells that begin producing ectopic cells will continue to do so, and that cells that initially produce normal cells also will continue to do so. Transitions between states appear to be very rare. Only 5 of 240 clusters from SE mice contained a mixture of ectopic and correctly located cells.

### Temporal/global regulation of basal dendrite formation

Basal dendrites were distributed close to the predicted ratio among clonal clusters and tended to be present in clusters in which the majority of cells lacked this feature. Progenitors that produce DGCs with basal dendrites, therefore, mostly produce morphologically normal DGCs. This observation suggests a mechanism that could impact the development of daughter cells from any progenitor, while also leaving most daughter cells unaffected. Such a mechanism might affect the entire hippocampus, but only some of the time. Seizure activity clearly meets these criteria, as seizures are episodic in nature, and when these seizures generalize, as is typical for the pilocarpine model, the entire hippocampus will be affected. Recent work by Botterill et al. (2015) supports this idea, with the demonstration that limbic kindling disrupts granule cell integration more severely than kindling of nonlimbic regions.

The idea that seizure activity might drive basal dendrite formation is supported by the work of Nakahara et al. (2009). They demonstrated that increasing neuronal activity in hippocampal slice cultures stabilized the normally transient basal dendrites that typically are present on immature granule cells. Under low-activity conditions, developing granule cells briefly possess basal dendrites 1 to 2 weeks after their birth, but subsequently reabsorb these processes as they mature. Increasing neuronal activity, however, allowed these processes to persist through cell maturity, perhaps through a neurotrophin-dependent mechanism (Danzer et al., 2002; Botterill et al., 2015). Whether a similar process occurs *in vivo* remains to be determined, but the present findings are consistent with the idea that episodic increases in activity (including seizures) might similarly stabilize granule cell basal dendrites, but only among granule cells that happen to be at this particular developmental stage at the time of the event.

### Depletion of the granule cell progenitor pool

In the current study, we found a decrease in progenitor cell numbers and an increase in differentiated cells. A chronic decrease in neurogenesis has been observed previously in epileptic animals (Hattiangady et al., 2004). Reduced neurogenesis could be the result of increased progenitor cell quiescence, loss of functional progenitor cell division, decreased survival of daughter cells, or an overall loss of progenitors. Our results provide evidence for the division-coupled loss of type 1 progenitor cells as a key contributor to the chronic decline in neurogenesis. We observed a decrease in quiescent and actively self-renewing progenitors, but an increase in mature granule cells within clonal groups in animals following status. Activation and terminal differentiation of quiescent progenitors would account for these observations. Indeed, Encinas et al. (2011) observed a similar loss of stem cells during the normal aging process in the mouse hippocampus. Using a genetic label, they showed that type 1 cells act as "disposable stem cells": once activated, they tend to terminally differentiate rather than continuing to cycle. Therefore, epileptic stimuli might accelerate the age-related loss of progenitor cells from the dentate. Sierra et al. (2015) observed a similar reduction in the progenitor cell pool following intrahippocampal injection of the convulsant kainic acid. In contrast to the present results, however, they observed terminal differentiation of type 1 cells into astrocytes (see also Hattiangady and Shetty, 2010), rather than mature granule cells, as described here. This difference likely reflects the very different pathological responses, and impacts on neurogenesis, of the two epilepsy models (Murphy et al., 2012).

### Concluding remarks

Our results strongly suggest different mechanistic origins for ectopic DGCs and DGCs with basal dendrites. Ectopic DGCs are highly localized to specific clonal clusters, implicating the parent progenitor cell or the neurogenic niche in which the progenitor resides. By contrast, basal dendrites appeared to be randomly distributed among clones, suggesting that transient changes acting throughout the hippocampus drive this pathology. Both abnormalities are implicated in the development of epilepsy and associated comorbidities, and separate therapeutic strategies will likely be required to mitigate these different abnormalities.